\documentclass[conference]{IEEEtran}
\IEEEoverridecommandlockouts
\usepackage{cite}
\usepackage{amsmath,amssymb,amsfonts}
\usepackage{graphicx}
\usepackage{textcomp}
\usepackage[table]{xcolor}

\usepackage{booktabs}
\usepackage{multirow}
\usepackage{float}
\usepackage{enumitem}
\usepackage{placeins}
\usepackage{eufrak}
\usepackage{tikz}

\usepackage{framed}
\usepackage{caption}
\usepackage{subcaption}
\usepackage{graphicx,xcolor}
\usepackage[framemethod=tikz]{mdframed}
\usepackage{algorithm}
\usepackage{algpseudocode}
\usepackage{lipsum}
\usepackage{url}
\usepackage{tikz}
\usepackage{array}
\usepackage{float}
\usepackage{soul}
\usepackage{hyperref}

\def\BibTeX{{\rm B\kern-.05em{\sc i\kern-.025em b}\kern-.08em
    T\kern-.1667em\lower.7ex\hbox{E}\kern-.125emX}}

\graphicspath{ {./img/} }
\definecolor{codegreen}{rgb}{0,0.6,0}
\definecolor{codegray}{rgb}{0.5,0.5,0.5}
\definecolor{codepurple}{rgb}{0.58,0,0.82}
\definecolor{backcolour}{rgb}{0.95,0.95,0.92}

\usepackage{listings}
\lstdefinestyle{mystyle}{
    backgroundcolor=\color{backcolour},
    commentstyle=\color{codegreen},
    keywordstyle=\color{magenta},
    stringstyle=\color{codepurple},
    basicstyle=\ttfamily\scriptsize,
    breakatwhitespace=false,
    breaklines=true,
    captionpos=b,
    keepspaces=true,
    showspaces=false,
    showstringspaces=false,
    showtabs=false,
    tabsize=2
}
\newcommand{\fname}{\emph{ai3}}

\makeatletter
\def\ps@IEEEtitlepagestyle{
  \def\@oddfoot{\mycopyrightnotice}
  \def\@evenfoot{}
}
\def\mycopyrightnotice{
  {\footnotesize 979-8-3503-8713-1/24/\$31.00~\copyright~2024 IEEE\hfill}
  \gdef\mycopyrightnotice{}
}

\makeatother
\usepackage{eso-pic}
\newcommand\AtPageUpperMyright[1]{\AtPageUpperLeft{
 \put(\LenToUnit{0.5\paperwidth},\LenToUnit{-1cm}){
     \parbox{0.5\textwidth}{\raggedleft\fontsize{9}{11}\selectfont #1}}
 }}
\newcommand{\conf}[1]{
\AddToShipoutPictureBG*{
\AtPageUpperMyright{#1}
}
}

\title{A Framework to Enable Algorithmic Design Choice Exploration in \emph{DNNs}}
\conf{IEEE High Performance Extreme Computing Conference (HPEC) 23-27 September, 2024}

\begin{document}

\author{\IEEEauthorblockN{Timothy L. Cronin IV, Sanmukh Kuppannagari}
\IEEEauthorblockA{\textit{Department of Computer and Data Sciences, Case Western Reserve University}\\
\textit{Contact: \{tlc107, sxk1942\}@case.edu}}}

\makeatletter
\patchcmd{\@maketitle}
  {\addvspace{0.5\baselineskip}\egroup}
  {\addvspace{-1\baselineskip}\egroup}
  {}
  {}
\makeatother
\maketitle

\begin{abstract}
Deep learning technologies, particularly deep neural networks (\emph{DNNs}), have demonstrated significant success across many domains. This success has been accompanied by substantial advancements and innovations in the algorithms behind the operations required by \emph{DNNs}. These enhanced algorithms hold the potential to greatly increase the performance of \emph{DNNs}. However, discovering the best performing algorithm for a \emph{DNN} and altering the \emph{DNN} to use such algorithm is a difficult and time consuming task. To address this, we introduce an open source framework which provides easy to use fine grain algorithmic control for \emph{DNNs}, enabling algorithmic exploration and selection. Along with built-in high performance implementations of common deep learning operations, the framework enables users to implement and select their own algorithms to be utilized by the \emph{DNN}. The framework's built-in accelerated implementations are shown to yield outputs equivalent to and exhibit similar performance as implementations in \emph{PyTorch}, a popular \emph{DNN} framework. Moreover, the framework incurs no additional performance overhead, meaning that performance depends solely on the algorithms chosen by the user.
\end{abstract}

\begin{IEEEkeywords}
Deep Neural Network, Algorithmic Design Choices, PyTorch
\end{IEEEkeywords}

\section{Introduction}

Artificial Intelligence (\emph{AI}) and more specifically deep learning which utilizes deep neural networks (\emph{DNNs}), has had a profound impact on society and is now a core part of modern technology. Due to its ability to achieve significant success in many classification, regression and generative problems, deep learning has seen applications in many fields\cite{dl_overview}. Fields such as healthcare\cite{dl_healthcare}, computer vision\cite{dl_compvis}, language\cite{dl_lang}, speech recognition\cite{dl_speech_recog}, cybersecurity\cite{dll_cybersec}, and many more.

Due to this impact, extensive research is being conducted in hopes of improving \emph{DNNs'} performance. Efforts to achieve these improvements span advancements in algorithms, hardware, architectures and more\cite{survey_on_improvements}. As the number of parameters in state-of-the-art \emph{DNN} models grows to reach one billion in the field of computer vision\cite{dinov2} and far surpass one billion in the field of large language models\cite{chatgpt3}, enhancing performance becomes imperative.

Algorithmic innovations hold the potential to significantly enhance the efficiency of \emph{DNNs}.  For example, from $2012-2019$ algorithmic innovations lead to doubling the efficiency of computer vision models every $16$ months, for large language models algorithmic innovations lead to doubling efficiency every $6$ months over a course of $3$ years\cite{measuring_algorithms}. These improvements can be achieved by reducing the number of computations, reducing memory usage and other methods.

Convolution, which is the backbone of many computer vision models, namely convolutional neural networks (\emph{CNNs})\cite{cnn}, has had several algorithms developed to perform the operation. Similarly, attention\cite{attention} which is the operation behind transformer models and many large language models\cite{attention_all_need} has seen several algorithms developed for it. Selecting the appropriate algorithm, though, is not a simple task. It depends on the use case of the \emph{DNN} and various other factors, such as input size, operation configuration, and available hardware.

In addition, incorporating different algorithms for the operations used in a \emph{DNN} is a non-trivial task. A task which requires expertise in both the algorithms and the format the \emph{DNN} exists in. Thus, the ability to easily explore the impact of algorithms and select the algorithms in use by a \emph{DNN} has been lost.

To address this, in this work, we present an open source framework\footnotemark[1] that enables a user to easily select the algorithms used in the operations of a \emph{DNN}, facilitating the exploration of various algorithmic design choices and potentially leading to a more efficient \emph{DNN}. The framework is titled \fname, an abbreviation for ``algorithmic innovations for accelerated implementations of artificial intelligence''. \fname\ provides accelerated \emph{C++}\cite{cpp} implementations of various algorithms for deep learning operations as well as a frontend in \emph{Python}\cite{python} which gives users fine grain control over the algorithms used by their \emph{PyTorch} \emph{DNN}. \fname\ also allows users to implement their own algorithms in \emph{C++} and easily include these algorithms when installing the package. After installation, the user's implementations can be selected in the same manner the built-in implementations are. The algorithmic selection and ability for users to implement their own algorithms provides the possibility of algorithmic innovations.

\footnotetext[1]{Source code at \emph{github.com/KLab-AI3/ai3}}
\section{Background}
\subsection{Deep Neural Networks}

Deep neural networks are the primary method used to apply deep learning technologies\cite{dl_overview}. Deep neural networks chain various layers in sequence to form a single computational model. The different layers represent the various operations performed on the data. Deep neural networks typically include many hidden layers along with the input and output layers. Input layers receive the data and output layers form the output from the altered data they receive. The hidden layers between the input and output layers are the networks computational engine.

Usually, \emph{DNN} development is split into two stages. The first stage is the training stage where the parameters of the operations performed to the data are learned. The strategy for learning the parameters can vary greatly between \emph{DNNs}, depending on the type of output the model produces, the data available, the operations performed in the model and more. The second stage is the inference stage in which the capabilities developed during the training stage are used to produce outputs. In this stage the \emph{DNN} infers an output, based on its training, from data it has not encountered before.

\subsection{Algorithmic Design Choices}

Fine grain levels of control over the algorithms used in the \emph{DNNs} operations can yield great benefits in performance. For example, changing the algorithm which performs the convolution in \emph{CNNs} can alter the latency and memory usage of the model greatly\cite{kn2row}\cite{better_direct}. These results are reflected in transformer models as well\cite{linformer}. Despite the existence of numerous algorithms, there is no single best algorithm for a given operation. Selecting the appropriate one is not a trivial task and depends on the specific \emph{DNN} use case, memory and time constraints, available hardware, and other factors\cite{on_nn_speed}.

Many different algorithms to perform the convolution operation have been developed, these include \emph{IM2COL}, \emph{KN2ROW}\cite{kn2row}, scalar \emph{MM}\cite{smm_conv} and more.

\begin{enumerate}[label={(\alph*)}]
\item \emph{IM2COL:} The image to column technique transforms the convolution operation to a matrix multiplication operation by reshaping the input and kernel to column vectors. Once this is complete, highly optimized general matrix multiply (\emph{GEMM}) routines can be utilized.
\item \emph{KN2ROW:} The kernel to row technique is also used to transform the convolution to matrix multiplication but differs in that it transforms the kernel into row vectors in order to decrease memory usage. After transforming the problem to one of matrix multiplication the same highly optimized routines can be utilized.
\item \emph{SMM:} Scalar Matrix Multiplication with Zero Packing avoids matrix multiplications and replaces it with matrix scaling operations. Each output image can be considered as the summation of shifted versions of the input image multiplied by the corresponding kernel weight.
\item \emph{Direct:} Direct convolution, as the name suggests, is a straight forward approach to convolution. Kernels are applied directly to the input without transforming the data.
\end{enumerate}

The ability to choose between these algorithms is key to enhancing a \emph{CNNs} performance as \emph{CNNs} spend the vast majority of their time inferencing on performing the convolutions\cite{time_in_conv}. Determining the best algorithm for a convolution operation is a non-trivial task depending on many features of the operation such as input sizes, number of output channels, kernel dimensions, stride of the kernel and more\cite{dif_algo_times_conv}.

Similar to \emph{CNN}s, transformer models spend the vast majority of their inferencing time in utilizing one mechanism, attention\cite{attention_all_need}. Attention is an operation used in \emph{DNNs} for language processing, it is used as it provides the benefit of being able to predict the target word depending on the context associated with the source, the \emph{DNN} is aware of what part of the source should receive the most attention. In response to this, many different algorithms have been developed in order to complete the operation. These include \emph{flash}, \emph{bigbird}, \emph{longformer}, \emph{linformer} attention and more.

\begin{enumerate}[label={(\alph*)}]
\item \emph{Flash:} An exact attention algorithm that uses tiling to increase the number of cache hits and decrease the number of memory read and writes between \emph{GPU} high bandwidth memory. 
\item \emph{LongFormer:} A self-attention algorithm which scales linearly as opposed to quadradically with input sequence length. This is achieved by combining a smaller windowed attention mechanism with a larger global attention mechanism.
\item \emph{BigBird:} A sparse attention mechanism which also reduces the quadratic dependency on input size to linear. BigBird is a universal approximator of sequence to sequence functions which preserves the properties of the quadratic full attention models.
\item \emph{Linformer:} An algorithm that approximates the attention mechanism reducing the time and space complexity from quadratic to linear. The algorithm has been found to perform on par with standard attention implementations.
\end{enumerate}

Again, the algorithm providing the best performance depends on many factors such as, input sequence length, memory limitations and required accuracy\cite{comp_attn}.

\subsection{Relevant Related Works}
One library that has many implementations of deep learning operations is the \emph{NVIDIA} \emph{CUDA} Deep Neural Network library \emph{(cuDNN)}\cite{cuDNN}. \emph{cuDNN} provides highly tuned, \emph{GPU} accelerated implementations of common routines utilized in deep learning. These implementations include forward and backward operations for convolution, \emph{matmul}, attention, pooling and normalization. The \emph{cuDNN} library solely exposes a \emph{C}\cite{c} \emph{API} meaning that it is non-trivial to explore different algorithms and select the algorithms in use by a \emph{DNN}.

\emph{PyTorch} is a leading \emph{Python} package used for training and inferencing with \emph{DNNs}. It is a highly optimized framework, especially for \emph{GPUs}. When using \emph{PyTorch's} implementations, however, there is currently no support for selecting the algorithm used to complete the operation. \emph{PyTorch} uses \emph{cuDNN} when possible and if \emph{torch.backends.cudnn.benchmark} is set to true will try to discover and use the fastest implementation from \emph{cuDNN}. However, support for manually setting the algorithm from \emph{cuDNN} to use is not supported, though it may be in the future.

One package which seeks to optimize \emph{PyTorch} models is the Intel Extension for \emph{PyTorch} \emph{(IPEX)}\cite{ipex}. \emph{IPEX} provides a very easy way to optimize \emph{PyTorch} models via a \emph{Python} front end to their \emph{C++} implementations. Though this enables users to easily optimize their models, given they have the proper hardware, \emph{IPEX} does not allow users to customize the algorithms used in the optimized models.
\section{Framework Overview}

The framework, titled \fname, enables users to seamlessly select different algorithms to be used in each layer of a \emph{DNN}. Additionally, it provides a straightforward library which can be used to develop custom implementations of the operations used in the \emph{DNN}. These custom implementations are then compiled along with the built-in ones and can be selected for use in the same manner the built-in implementations are. \fname\ currently supports the following operations, linear, convolution, flatten, \emph{ReLU}, and adaptive average, max, and average pooling.

\begin{figure}[ht]
    \centering
    \captionsetup{justification=centering}
    \includegraphics[width=\linewidth]{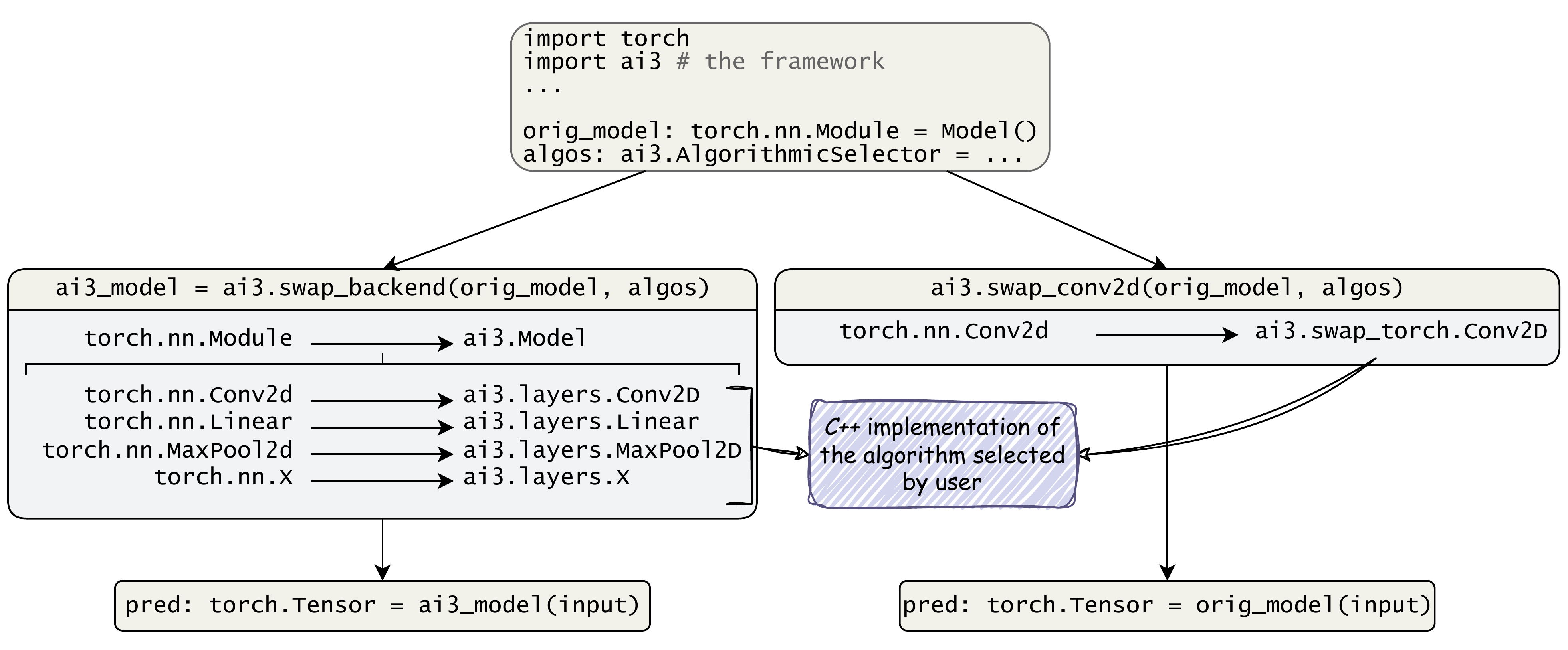}
    \caption{Example paths of a \emph{PyTorch} module through the framework\protect\footnotemark[2]}
\end{figure}
\footnotetext[2]{Created with \emph{draw.io}\cite{drawio}}

\subsection{Use Cases}

\subsubsection{\emph{DNN} Application Development}

The first use case \fname\ seeks to address is a user who does not wish to implement their own algorithms but still wishes to customize the algorithms the existing \emph{DNN} uses. This user is expected to be able to use \emph{Python} and \emph{PyTorch} but does not need to be familiar with \emph{C++} or accelerated computing. For these users all that must be done to enable algorithmic selection is importing the package and adding a single line of \emph{Python} code calling one of two functions which perform algorithmic selection on the \emph{PyTorch} \emph{DNN}. 

\subsubsection{\emph{DNN} Algorithm Development}

The second use case \fname\ seeks to address is a user developing implementations of custom algorithms for \emph{DNNs}. For these users, \fname\ provides a \emph{C++} library containing a \emph{Tensor} class and various utility functions to do complete simple tasks and improve the developer experience. \fname\ also provides \emph{C++} placeholder functions where users should implement their algorithms. After the package is installed, users can select their custom implementations for use by the \emph{DNN}.

\section{Framework Implementation}

Depending on the use case, there are two paths for installing the package. For the first use case, \fname\ is available as a \emph{Python} package that can be installed via \emph{pip}, the package installer for \emph{Python}, with the command, \emph{pip install aithree}. During installation, supported parallel computing platforms and libraries will be searched for and if found, the built-in implementations using such platforms and libraries will be utilized. For the second use case, where a user wishes to implement algorithms manually, users must download the source code instead of installing with \emph{pip}. After downloading, users will find a \emph{C++} header file for each operation. The header file contains the function signature of the operation to be implemented, access to the framework's \emph{C++} library, and a boolean which controls default algorithm selection using the custom algorithm. As the installation process is able to compile arbitrary \emph{C++} code, any existing \emph{C++} compatible library can be used for the users implementation. Therefore, the ability to use existing libraries like \emph{cuDNN}, \emph{oneMKL}\cite{onemkl}, etc. is not lost. Operability between \emph{Python} and \emph{C++} is provided via \emph{PyBind11}\cite{pybind11}. The framework utilizes a build system consisting of \emph{scikit-build-core}\cite{scikit-build-core} and \emph{CMAKE}\cite{cmake}. After implementing their algorithms, users can install the package using \emph{pip} with the installation target pointing to their local directory, again, parallel computing platforms and libraries on the machine will be utilized if possible. After installing the package, \fname\ provides a comprehensive test suite which can be used to ensure the correctness of custom implementations. 

After installing the package, two functions to perform algorithmic selection become available for use in \emph{Python}.  The first, \emph{swap\_backend}, replaces all operations in a \emph{DNN} with \fname's implementation of the operation using the selected algorithm, resulting in a \emph{DNN} fully managed by \fname. The second, \emph{swap\_(module type)}, swaps out all operations of a specific type with \fname's implementation of the operation using the chosen algorithm. Code sample~\ref{lst:basic} demonstrates use of these functions.

\begin{lstlisting}[language=Python, caption={Basic Use of the Framework}, label={lst:basic}]
import torch
from torch import nn
import ai3  # the framework

class ConvNet(nn.Module):
    def __init__(self):
        super(ConvNet, self).__init__()
        self.conv1 = nn.Conv2d(in_channels=3,
                               out_channels=16,
                               kernel_size=3, padding=1)
        self.maxpool = nn.MaxPool2d(kernel_size=2,
                                    stride=2)
        self.conv2 = nn.Conv2d(in_channels=16,
                               out_channels=32,
                               kernel_size=3, padding=1)

    def forward(self, x):
        x = torch.relu(self.conv1(x))
        x = self.maxpool(x)
        x = torch.relu(self.conv2(x))
        x = torch.flatten(x, 1)
        return x

input_data = torch.randn(10, 3, 224, 224)
orig = ConvNet()
torch_out = orig(input_data)
model: ai3.Model = ai3.swap_backend(orig,
                                    {"conv2d": "direct"})
sb_out = model(input_data)
ai3.swap_conv2d(orig, ["direct", "smm"])
sc_out = orig(input_data)
assert torch.allclose(torch_out, sb_out, atol=1e-6)
assert torch.allclose(torch_out, sc_out, atol=1e-6)
\end{lstlisting}

The first function called, \emph{swap\_backend}, replaces every \emph{PyTorch} module and function used with \fname's implementation. It returns an object completely managed by the framework. The second function, \emph{swap\_conv2d}, replaces, in place, every instance of a \emph{PyTorch} \emph{torch.nn.Conv2d}, with another \emph{torch.nn.Module} that uses \fname's implementation of the selected algorithm. Both of these functions enable fine grain customization of the algorithms in use by the \emph{DNN}. When swapping, each module type is associated with one of three algorithmic selectors. The simplest is a single string containing the name of algorithm to use for all modules of that type. In code sample~\ref{lst:basic}, this strategy is used in the call to \emph{swap\_backend}, meaning that all the convolution layers in the \emph{ai3.Model} utilize direct convolution as \emph{``direct''} was passed. Next, a list of algorithms to use can be passed. As modules are encountered, they are replaced with an implementation of the algorithm in the list with the same index as that module has relative to the other modules of the same type. In code sample~\ref{lst:basic}, this strategy is used in the call to \emph{swap\_conv2d}, meaning the first convolution module is replaced with an implementation of direct convolution and the second convolution module is replaced with an implementation of \emph{SMM} convolution. The third method for algorithmic selection is a function which returns the algorithm to use and whose single parameter is the module the framework is currently replacing. This strategy enables powerful automated algorithmic selection. Code sample~\ref{lst:func_selector} demonstrates this strategy.
\begin{lstlisting}[language=Python, caption={Sample Using a Function to Select the Algorithms}, label={lst:func_selector}]
import torch
import torchvision
import ai3 # the framework

def selector(orig: torch.nn.Conv2d) -> str:
    in_channels = orig.weight.shape[1]
    if in_channels > 200:
        return "smm"
    return "direct"

input_data = torch.randn(1, 3, 224, 224)
vgg16 = torchvision.models.vgg16(
    weights=torchvision.models.VGG16_Weights.DEFAULT)
vgg16.eval()
torch_out = vgg16(input_data)

model: ai3.Model = ai3.swap_backend(vgg16,
                                    {"conv2d": selector})
sb_out = model(input_data)
assert torch.allclose(torch_out, sb_out, atol=1e-4)

ai3.swap_conv2d(vgg16, selector)
sc_out = vgg16(input_data)
assert torch.allclose(torch_out, sc_out, atol=1e-4)
\end{lstlisting}

When using \emph{swap\_backend}, a mapping is passed from the names of the modules to one of the three algorithmic selectors. If the user defines custom algorithms to use, they could select them explicitly by passing \emph{``custom''} for the algorithm name. If the user has not made selections on which algorithms to use for a module, equivalent to passing \emph{``default''} as the algorithm, then one of the framework's implementations is selected unless there is a user defined implementation for that operation in which case that is used. Additionally, if the user is swapping algorithms in a \emph{PyTorch} \emph{DNN} in place using \emph{swap\_(module type} then a \emph{``torch''} can be passed to keep the current \emph{PyTorch} implementation.

As shown in samples~\ref{lst:basic} and~\ref{lst:func_selector}, after calling the functions the altered object can be used to make predictions the same way a \emph{PyTorch} \emph{DNN} is.

In order to properly construct an equivalent \emph{DNN}, a symbolic pass through the original \emph{DNN} must be performed, storing all the operations encountered in a graph. This symbolic trace is achieved using the \emph{PyTorch} function, \emph{torch.fx.symbolic\_trace}, this function performs the pass and returns a graph. This graph is then searched through to either construct a module separate from \emph{PyTorch} in the case of a \emph{swap\_backend} call, or alter the \emph{PyTorch} \emph{DNN} in the case of a \emph{swap\_(module type)} call.

In order to construct a \emph{DNN} completely managed by the framework, instances of the framework's modules are created and stored in a list in the order they are reached in the pass. This list is then passed to another module which constructs a single callable, an \emph{ai3.Model}, containing all of them. As the list is formed, each module is given the algorithm the user chose to ensure that it utilizes the proper algorithm. The module which is returned to the user is completely managed by the framework and is able to perform all of its operations in \emph{C++} without any \emph{Python} until it has completed its operations. 

On the other hand replacing only some operations of a \emph{PyTorch} module requires that the replacements are descendants of the \emph{torch.nn.Module} class. The replacement \emph{torch.nn.Modules} act as slim wrappers around the frameworks equivalent operation whose forward function calls the implementation of the algorithm selected by the user.

As noted earlier, the framework provides an easy to use and robust testing suite. First, unit tests are employed to validate each operation. By varying hyperparameters, input sizes and more, we ensure that the operations remain equivalent to \emph{PyTorch's} regardless of the operation's configuration. For example, when testing the \emph{MaxPool2D} implementation against \emph{PyTorch's} implementation, parameters such as kernel size, padding, dilation and stride are varied. Additionally, the input shape is adjusted by modifying the batch size, number of channels, height, and width. In addition to the unit tests, tests for both the \emph{swap\_backend} and \emph{swap\_(module type)} functions are present. These functions are performed by first creating a \emph{PyTorch} \emph{DNN}, then performing the swap and then performing a forward pass on the same input data, ensuring the results are the same. The original \emph{PyTorch} modules used in these tests include various models in the \emph{torchvision}\cite{torchvision2016} package along with some manually created \emph{PyTorch} modules. For all tests, including unit tests, if more than one algorithm is available, all algorithms implemented will be tested to ensure the correctness of every implementation.
\section{Experimental Evaluations}

\subsection{Setup}
Experiments and evaluations are performed on the High Performance Computing Resource in the Core Facility for Advanced Research Computing at Case Western Reserve University. The operations are performed on an \emph{NVIDIA GeForce RTX} $2080$ \emph{Ti} \emph{GPU}. Benchmarks are conducted on a single convolutional module as well as many prevalent \emph{CNNs}. The \emph{CNNs} used for benchmarks are \emph{AlexNet}\cite{alexnet}, \emph{DenseNet}\cite{densenet}, \emph{GoogleNet}\cite{googlenet}, Inception-\emph{v$3$}\cite{inceptionv3}, \emph{ResNet}\cite{resnet}, \emph{Swin} Transformer\cite{swintransformer}, Vision Transformer\cite{visiontransformer}, \emph{Squeezenet}\cite{squeezenet} and \emph{VGG$16$}\cite{vgg16}. The input shapes used for the single convolution module are $(10, 3, 224, 224)$, $(10, 64, 112, 112)$, $(10, 256, 28, 28)$, and $(10, 512, 14, 14)$. The input shape used for all the \emph{CNNs} is $(10, 3, 224, 224)$. The dimensions of the input correspond to $(\text{batch size}, \text{channels}, \text{height}, \text{width})$.

When benchmarking, we perform the operation using \emph{PyTorch} in eager mode, along with multiple algorithms implemented by the framework. The benchmarks focus on convolution as it is a computationally expensive operation and acts as the core of many prevalent \emph{DNNs}.

\subsection{Algorithm Implementation}

When performing benchmarks on the \emph{GPU}, \emph{PyTorch} makes use of \emph{cuDNN} for implementations. Algorithms implemented by the framework and used in the benchmarks also rely on \emph{cuDNN} to provide times comparable with \emph{PyTorch} and demonstrate the low overhead of \fname. Operations are performed within a \emph{torch.inference\_mode()} context to avoid any computational overhead associated with preparing for gradient calculation. The algorithms implemented by \emph{cuDNN} and used by \fname\ while performing benchmarks are, implicit \emph{precomp} \emph{GEMM}, implicit \emph{GEMM}, \emph{Winograd}\cite{winograd}, \emph{GEMM} and using the algorithm selected by the \emph{cuDNN} function \emph{cudnnGetConvolutionForwardAlgorithm\_v7}, which serves as a heuristic for obtaining the best suited algorithm.
\begin{enumerate}[label={(\alph*)}]
\item Implicit \emph{precomp} \emph{GEMM}: Expresses the convolution as a matrix product while not actually forming the necessary matrix. Various indices are calculated in a precomputation step which requires memory but assists the operation.
\item Implicit \emph{GEMM}: A zero memory overhead algorithm which expresses the convolution as a matrix product while not actually forming the necessary matrix.
\item \emph{GEMM}: Expresses the convolution as a matrix product and explicitly forms the matrix necessary.
\item \emph{Winograd}: Utilizes results of minimal complexity convolutions to increase performance when using smaller filters and batch sizes.
\end{enumerate}

\subsection{Results}

\begin{figure}[ht]
    \centering
    \captionsetup{justification=centering}
    \includegraphics[width=\linewidth]{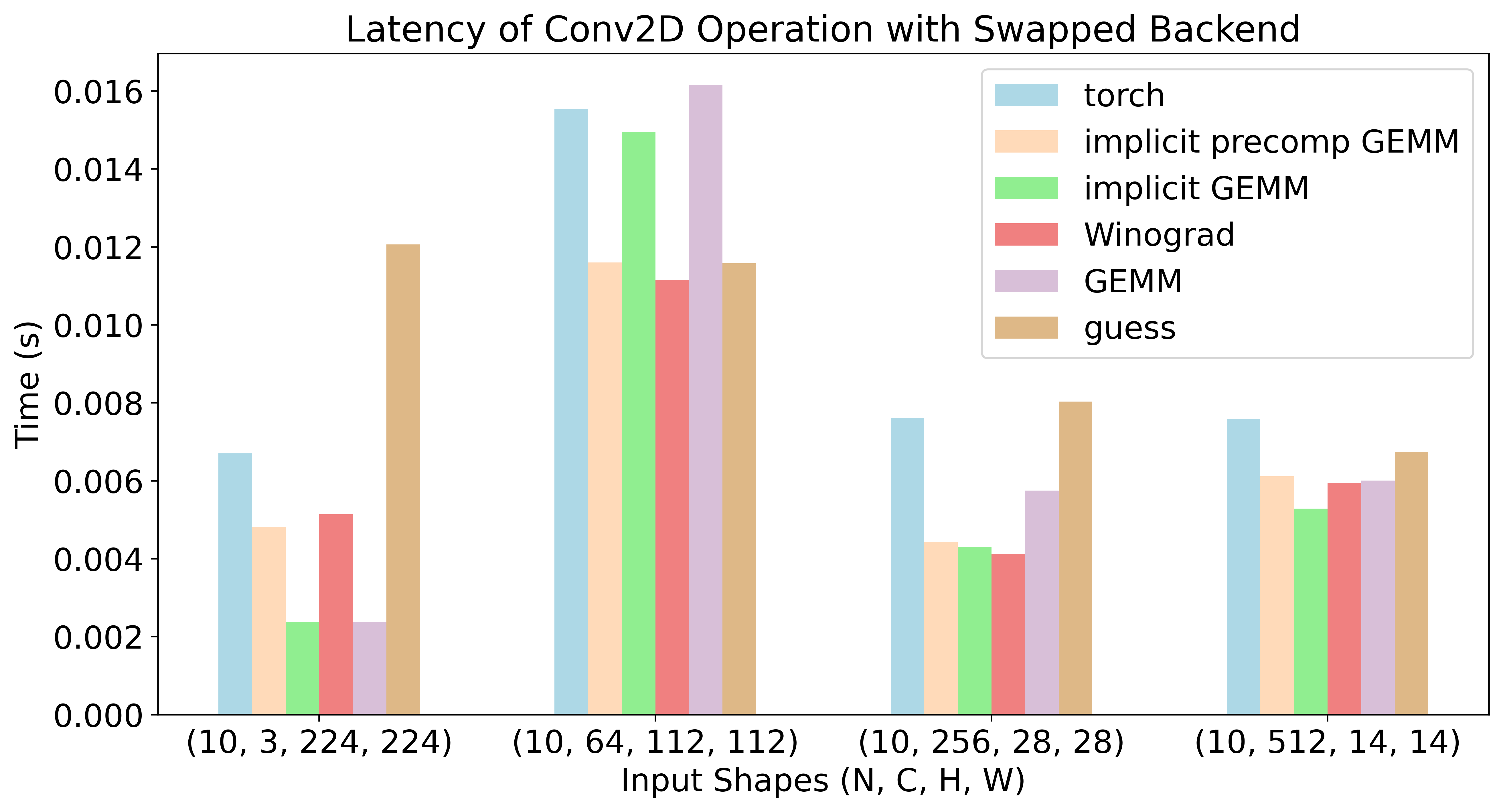}
    \caption{Latency of Algorithms Using Swapped Backend}
    \label{fig:swap_backend}
\end{figure}

\begin{figure}[ht]
    \centering
    \captionsetup{justification=centering}
    \includegraphics[width=\linewidth]{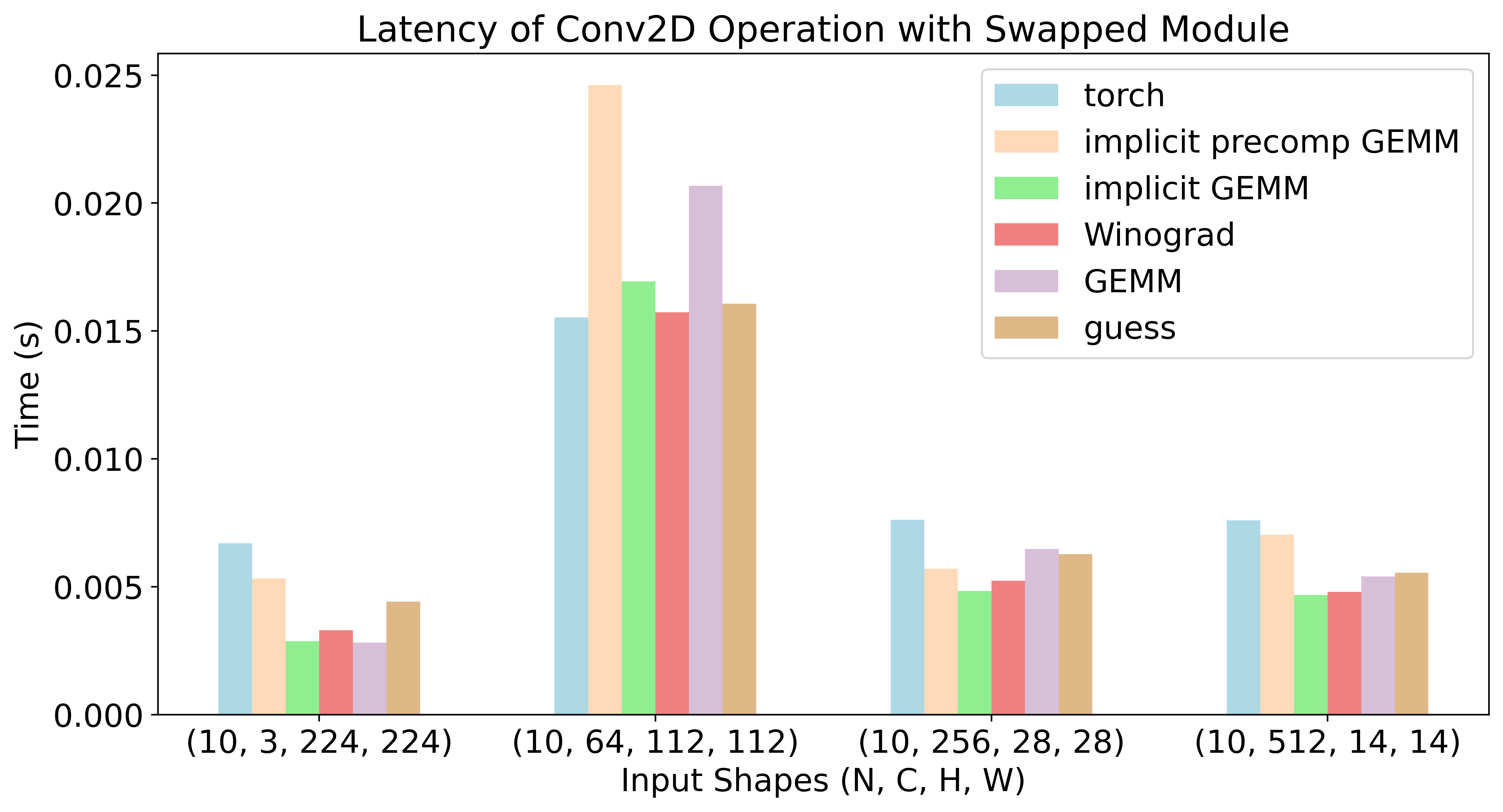}
    \caption{Latency of Algorithms Using Swapped Modules}
    \label{fig:swap_module}
\end{figure}

Figures~\ref{fig:swap_backend} and~\ref{fig:swap_module} represent the latency of a single convolution operation. The different colored bars represent the different algorithms. All algorithms but \emph{torch} are run utilizing \fname. The \emph{guess} algorithm uses the algorithm given by the \emph{cudnnGetConvolutionForwardAlgorithm\_v7} function described earlier. Figure~\ref{fig:swap_backend} shows latency where algorithmic selection was performed via a call to the \emph{swap\_backend} function.

In contrast, in figure~\ref{fig:swap_module}, algorithmic selection was performed via a call to the \emph{swap\_conv2d} function.

\begin{figure}[ht]
    \centering
    \captionsetup{justification=centering}
    \includegraphics[width=\linewidth]{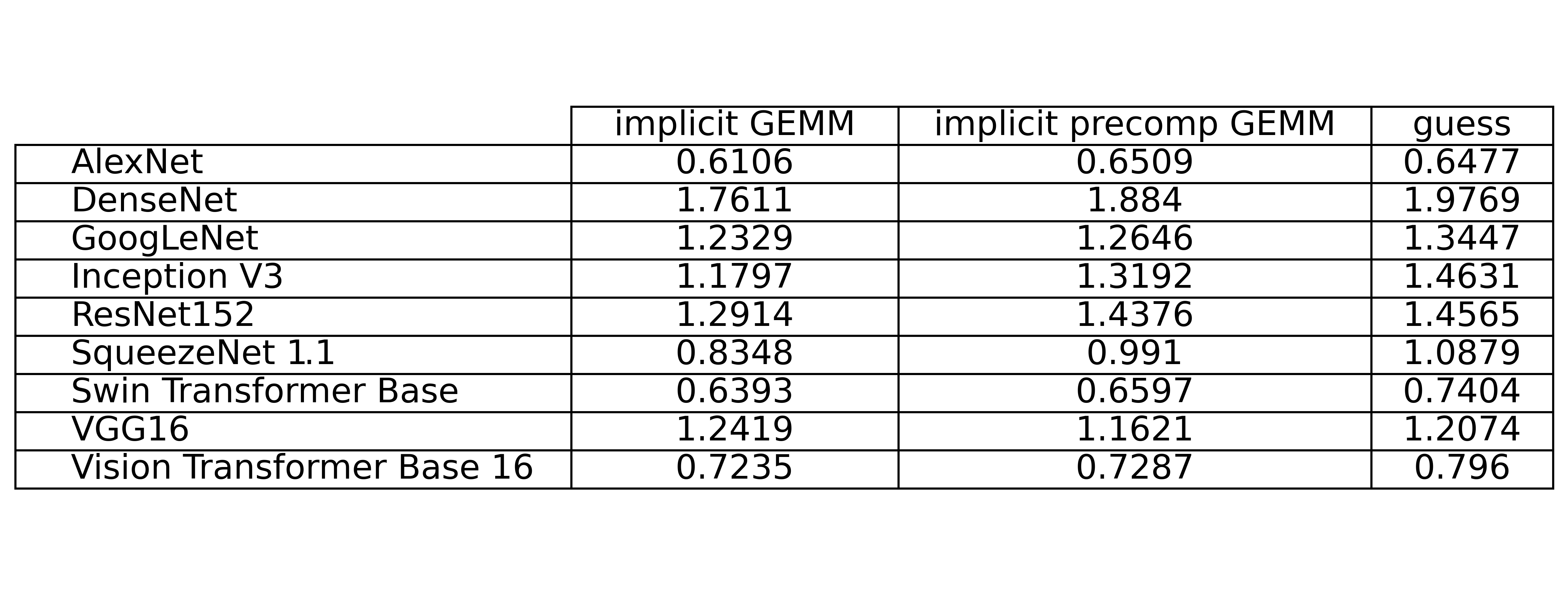}
    \caption{End-to-End Latency of \emph{CNNs} with Swapped Modules Relative to \emph{PyTorch}}
    \label{fig:model_times}
\end{figure}

Figure~\ref{fig:model_times} describes the end-to-end latency of the \emph{CNNs} listed earlier processing $10$ samples. The \emph{PyTorch} convolution modules are swapped out for \fname's implementation of the given algorithm via \fname's \emph{swap\_conv2d} function. The elements are normalized per row with respect to the latency of \emph{PyTorch}. This means that algorithms with values $<1$ have lower latency and algorithms with values $>1$ have greater latency than \emph{PyTorch}.

Figures ~\ref{fig:swap_module} and ~\ref{fig:swap_backend} which describe the latency of a single convolution operation generally show lower latency when utilizing \fname\ instead of \emph{PyTorch}. As both packages utilize \emph{cuDNN} implementations to perform the convolution, these results suggest that the latency in starting and finalizing the operation is lower in \fname\ compared to \emph{PyTorch}, demonstrating the low overhead of \fname. The difference in latency between \fname\ and \emph{PyTorch} is generally larger on bar graph~\ref{fig:swap_backend} which shows the latencies of an \emph{ai3.Model} which is returned by \emph{swap\_backend}. This again demonstrates the low overhead of the framework as a \emph{ai3.Model} is generally able to commence the operations with lower latency than a \emph{PyTorch} \emph{nn.Module}. However, these results are not reflected in table~\ref{fig:model_times} which displays the relative latencies of the \emph{CNNs} listed earlier. On average, the original \emph{PyTorch} \emph{CNNs} have lower latency before swapping the convolution modules for \fname's implementations of the given algorithm. This is likely due to optimizations for successive convolution calls in \emph{PyTorch} that are not reflected in \fname.  For example, \emph{PyTorch} may be caching data for reuse to avoid reallocating and reconfiguring. Implementing such optimizations for the framework is possible and will be useful for improving performance.
\section{Limitations and Future Directions}

Currently, some popular modules in \emph{PyTorch} are still not implemented by \fname. This means that \emph{swap\_backend} is often not able to replace all operations of the \emph{PyTorch} \emph{DNN}. Adding support for additional modules will contribute towards \emph{swap\_backend} executing successfully in the majority of cases allowing an easy way to configure all algorithms utilized by the \emph{DNN}. Until then, repeated calls to \emph{swap\_(module type)} functions can be made to enable algorithmic selection on as many modules as possible.

Secondly, \fname\ does not have built-in support for grouped convolution. However, once implemented, grouped convolution will be trivial to integrate and users will not have to make any change to their code. \fname\ is still able to perform grouped convolution provided the user implements a convolution algorithm that has grouped support using the custom algorithm features of the framework described earlier.

Lastly, some operations do not have accelerated implementations. Once all operations have accelerated implementations, \emph{swap\_backend} should provide similar or improved performance compared to the original \emph{PyTorch} \emph{DNN}. Until this is achieved, users can swap out as many modules as desired for \fname's accelerated implementations using the \emph{swap\_(module type)} functions.

All of these limitations are not permanent and being worked on currently.

There are many future directions for \fname. First, support for more operations will be implemented. With the rise of transformer models, enabling algorithmic selection between various attention algorithms is becoming more and more crucial. Implementations for attention are being worked on and once implemented can be integrated into \fname\ and easily usable.

Second, in order to improve performance and integrate better with \emph{PyTorch}, the ability to use \fname while in \emph{PyTorch} graph mode is planned. This will enable improved performance as the operations surrounding \fname's will be optimized. It is also planned to provide similar optimizations as \emph{torch.compile} when using the \emph{swap\_backend} function.

Third, to enhance \fname's portability, support for other \emph{DNN} representations like the \emph{ONNX}\cite{onnx} format and \emph{TensorFlow}\cite{tensorflow} is planned. This will enable broader algorithmic selection and accessibility for a wider range of \emph{DNNs} and users.

Lastly, backward propagation is planned to be supported. Integrating these computations into various \emph{DNN} frameworks may be challenging, but it will enable powerful algorithmic selection for both the forward and backward passes of the \emph{DNN}. Meaning users will have fine grain algorithmic control throughout the \emph{DNN's} entire lifetime.

\section*{Acknowledgements}

This work was supported in part by a summer research award provided by \emph{CWRU} Undergraduate Research Office and by the U.S. National Science Foundation Award 2411447. Additionally, this work made use of the High Performance Computing Resource in the Core Facility for Advanced Research Computing at Case Western Reserve University. 

\section{Conclusion}
In this work we justified the existence of and demonstrated the usefulness of a framework that enables easy to use fine grain algorithmic selection for a \emph{DNN}. The framework, \fname, features a frontend in \emph{Python}, accelerated implementations of common deep learning operations, and a \emph{C++} library which can be used to implement custom algorithms using any valid \emph{C++} code. After installing the package two types of functions become available, one that swaps the entire backend of an existing \emph{DNN}, swapping all the modules for \fname's, and one that swaps a specific module type for \fname's implementation. These functions allow fine grain customization of the algorithms used by passing the names of the algorithms to use directly or passing a function which examines the module and returns the algorithm to use. These algorithms are benchmarked and \fname is shown to have low overhead. In fact, the result of a \emph{swap\_backend} call is often able to begin the computations required faster than alternatives. \fname\ provides algorithmic selection at low cost, bringing the possibility of higher performance.

\label{conclusion}

\bibliographystyle{IEEEtran}
\bibliography{0_conf}

\end{document}